\pdfoutput=1
\documentclass[11pt]{article} 
\usepackage[final]{acl}
\usepackage{times}
\usepackage{latexsym}
\usepackage[T1]{fontenc}
\usepackage[utf8]{inputenc}
\usepackage{microtype}
\usepackage{inconsolata}
\usepackage{graphicx}

\usepackage[utf8]{inputenc} 

\usepackage[T1]{fontenc}    
\usepackage{hyperref}       
\usepackage{url}            
\usepackage{booktabs}       
\usepackage{amsfonts}       
\usepackage{nicefrac}       
\usepackage{microtype}      
\usepackage{xcolor}         
\usepackage{standalone}
\usepackage{latexsym}
\usepackage{amsmath}
\usepackage{amssymb}
\usepackage{amsthm}
\usepackage{array}
\usepackage{tabu}
\usepackage{makecell}
\usepackage{paralist}
\usepackage{cases}
\usepackage{diagbox}
\usepackage{enumitem}
\usepackage{soul}
\usepackage{multirow}
\usepackage{verbatim}
\usepackage{tabulary}
\usepackage{booktabs}
\usepackage{tabularx}
\usepackage[mathscr]{euscript}
\usepackage{mathtools}
\usepackage{algorithm}
\usepackage{algpseudocode}
\usepackage{stmaryrd}
\usepackage{tikz-dependency}
\usetikzlibrary{automata,decorations.markings,arrows,positioning,matrix,calc,patterns,angles,quotes,calc}
\usepackage{adjustbox}
\usepackage{tabularx}
\usepackage{xspace}
\usepackage{tabulary}
\usepackage{afterpage}
\usepackage{bm}
\usepackage{color}
\usepackage{graphicx}
\usepackage{slashbox}
\usepackage[toc,page]{appendix}
\usepackage{makecell}
\usepackage{boldline}
\usepackage[shortcuts]{extdash}  

\usepackage{blindtext}
\usepackage{graphicx}
\usepackage{capt-of}
\usepackage{booktabs}
\usepackage{varwidth}
\usepackage{pifont}
\usepackage{wrapfig}

\usepackage[utf8]{inputenc}

\DeclareFixedFont{\ttb}{T1}{txtt}{bx}{n}{12} 
\DeclareFixedFont{\ttm}{T1}{txtt}{m}{n}{12}  

\usepackage{color}
\definecolor{deepblue}{rgb}{0,0,0.5}
\definecolor{deepred}{rgb}{0.6,0,0}
\definecolor{deepgreen}{rgb}{0,0.5,0}

\usepackage{listings}

\usepackage{tcolorbox}
\tcbuselibrary{minted,breakable,xparse,skins}

\definecolor{bg}{gray}{0.95}
\DeclareTCBListing{mintedbox}{O{}m!O{}}{%
  breakable=true,
  listing engine=minted,
  listing only,
  minted language=#2,
  minted style=default,
  minted options={%
    linenos,
    gobble=0,
    breaklines=true,
    breakafter=,,
    fontsize=\small,
    numbersep=8pt,
    #1},
  boxsep=0pt,
  left skip=0pt,
  right skip=0pt,
  left=25pt,
  right=0pt,
  top=3pt,
  bottom=3pt,
  arc=5pt,
  leftrule=0pt,
  rightrule=0pt,
  bottomrule=2pt,
  toprule=2pt,
  colback=bg,
  colframe=orange!70,
  enhanced,
  overlay={%
    \begin{tcbclipinterior}
    \fill[orange!20!white] (frame.south west) rectangle ([xshift=20pt]frame.north west);
    \end{tcbclipinterior}},
  #3}

\definecolor{orange}{rgb}{1,0.5,0}
\definecolor{mdgreen}{rgb}{0.05,0.6,0.05}
\definecolor{mdblue}{rgb}{0,0,0.7}
\definecolor{dkblue}{rgb}{0,0,0.5}
\definecolor{dkgray}{rgb}{0.3,0.3,0.3}
\definecolor{slate}{rgb}{0.25,0.25,0.4}
\definecolor{gray}{rgb}{0.5,0.5,0.5}
\definecolor{ltgray}{rgb}{0.7,0.7,0.7}
\definecolor{purple}{rgb}{0.7,0,1.0}
\definecolor{lavender}{rgb}{0.65,0.55,1.0}

\definecolor{mypurple}{RGB}{111,61,121}
\definecolor{myblue}{RGB}{46,88,180}
\definecolor{myred}{RGB}{181,68,106}
\definecolor{myyellow}{RGB}{204,143,55}

\newcommand{\ensuretext}[1]{#1}

\newcommand{\draftcomment}[3]{\ensuretext{\textcolor{#3}{[#1 #2]}}}

\newcommand{\interalia}[1]{\citep[\emph{inter alia}]{#1}}
\renewcommand{\draftcomment}[3]{}  

\newcommand{\term}[1]{\textbf{#1}} 

\DeclareSymbolFont{extraup}{U}{zavm}{m}{n}
\DeclareMathSymbol{\vardiamond}{\mathalpha}{extraup}{87}

\newcolumntype{L}[1]{>{\raggedright\let\newline\\\arraybackslash\hspace{0pt}}m{#1}}
\newcolumntype{C}[1]{>{\centering\let\newline\\\arraybackslash\hspace{0pt}}m{#1}}
\newcolumntype{R}[1]{>{\raggedleft\let\newline\\\arraybackslash\hspace{0pt}}m{#1}}

\theoremstyle{definition}

\theoremstyle{remark}

\algrenewcommand{\algorithmiccomment}[1]{\leavevmode$\triangleright$ #1}

\setul{1pt}{.4pt}

\DeclareFixedFont{\ttb}{T1}{txtt}{bx}{n}{12} 
\DeclareFixedFont{\ttm}{T1}{txtt}{m}{n}{12}  

\usepackage[capitalize]{cleveref}

\title{Context Length Alone Hurts LLM Performance Despite Perfect Retrieval}

\author{\textbf{Yufeng Du \textsuperscript{1}\thanks{Equal contribution.}},
  \textbf{Minyang Tian \textsuperscript{1}\footnotemark[1]},
   \textbf{Srikanth Ronanki \textsuperscript{2}}, 
   \textbf{Subendhu Rongali \textsuperscript{2}}, \\
   \textbf{Sravan Bodapati \textsuperscript{2}}, 
   \textbf{Aram Galstyan \textsuperscript{2, 3}}, 
   \textbf{Azton Wells\textsuperscript{4}}, \\
   \textbf{Roy Schwartz\textsuperscript{5}}, 
   \textbf{Eliu A Huerta\textsuperscript{4, 6}}, 
   \textbf{Hao Peng\textsuperscript{1}}
   \\  
  \textsuperscript{1}University of Illinois at Urbana-Champaign, 
  \textsuperscript{2}Amazon.com Inc., \\
  \textsuperscript{3}USC Information Sciences Institute,
  \textsuperscript{4}Argonne National Laboratory, \\
\textsuperscript{5}The Hebrew University of Jerusalem,
\textsuperscript{6}University of Chicago \\
\small{ 
    \textbf{Correspondence:}\texttt{ \{\href{mailto:yufengd4@illinois.edu}{yufengd4}, \href{mailto:mtian8@illinois.edu}{mtian8}, \href{mailto:haopeng@illinois.edu}{haopeng}\}@illinois.edu
  } }
  }

\begin{document}
\maketitle

\begin{abstract}

Large language models (LLMs) often fail to scale their performance on long-context tasks performance in line with the context lengths they support. 
This gap is commonly attributed to retrieval failures---the models' inability to identify relevant information in the long inputs. 
Accordingly, recent efforts often focus on evaluating and improving LLMs’ retrieval performance: if retrieval is perfect, a model should, in principle, perform just as well on a long input as it does on a short one---or should it? 
This paper presents findings that the answer to this question may be negative.
Our systematic experiments across 5 open- and closed-source LLMs on math, question answering, and coding tasks reveal that, even when models can perfectly retrieve all relevant information, their performance still degrades substantially (13.9\%--85\%) as input length increases but remains  well within the models' claimed lengths.
This failure occurs even when the irrelevant tokens are replaced with minimally distracting  whitespace, and, more surprisingly, when they are all masked and the models are forced to attend \emph{only} to the relevant tokens.
A similar performance drop is observed when all relevant evidence is placed immediately before the question.
Our findings reveal a previously-unrealized limitation: the sheer length of the input alone can hurt LLM performance, independent of retrieval quality and \emph{without} any distraction. 
They motivate our simple, model-agnostic mitigation strategy that transforms a long-context task into a short-context one by prompting the model to recite the retrieved evidence before attempting to solve the problem.
On RULER, we observe a consistent improvement of GPT-4o up to 4\% on an already strong baseline.

\end{abstract}

\section{Introduction}\label{sec:intro}

\begin{figure*}[htbp]
    \centering
    \includegraphics[width=0.72\linewidth]{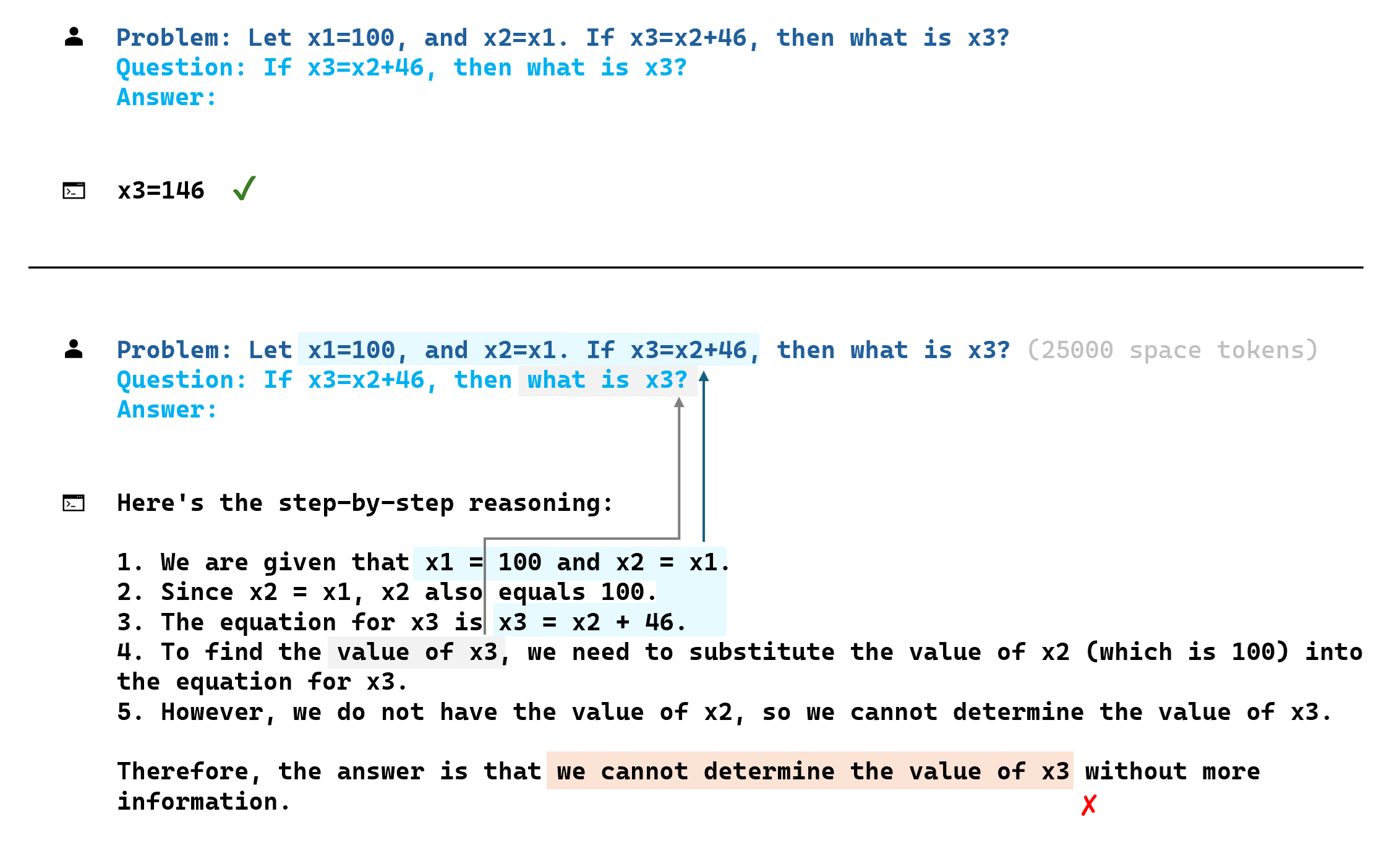}
    \caption{Extending the input length alone substantially degrades LLM reasoning capability, even if the model is still able to retrieve the relevant evidence. In this example, inserting 25000 white spaces (with minimal distraction) does not prevent the model from extracting all conditions and question correctly, but nevertheless causes it to reach the wrong answer. 
    }
    \label{fig:failure-to-solve-problem}
\end{figure*}

Recent large language models~(LLMs) have substantially expanded their context windows. For example, models like Llama-3 \cite{grattafiori2024llama3herdmodels} and Claude 3 \cite{anthropic2024claude3} can process 100K+ tokens, 
and Gemini can reportedly handle several million tokens~\cite{geminiteam2024gemini15unlockingmultimodal}.
The push to extend context windows of LLMs has raised expectations for their ability to solve problems over long inputs,
such as those that require integrating information across, e.g., multiple books, entire code repositories, or modeling long-horizon conversation \citep{Chang2023BooookScoreAS,Liu2024RepoQAEL,Stallone2024ScalingGC}. 
However, the growing capacity of LLMs to process long inputs has not consistently translated into a corresponding capability to effectively solve tasks over long contexts \interalia{hengle2025llmsreasonextendedmultilingual,Lee2024LongContext,kuratov2024babilong}.
What, then, prevents models from turning access to information into effective use for problem-solving over long contexts?

Recent studies suggest that LLMs approach long-context tasks through two interleaved processes: (1) identifying relevant information within the input (commonly referred to as \term{retrieval}\footnote{
Our use of the term retrieval follows standard practice in the long-context LLM literature \interalia{kamradt2023needleinahaystack,wu2024retrievalheadmechanisticallyexplains,hsieh2024ruler}. 
It refers to a model's ability to attend to and recite specific texts from the context, and should not be confused with retrieval in retrieval-augmented generation (RAG), which typically involves search engines.
}), and (2) using it to solve the problem~\interalia{wu2024retrievalheadmechanisticallyexplains,kuratov2024babilong,li2024alr2retrievethenreasonframeworklongcontext,zhang2025attentionrevealstokenstrainingfree}.
This conceptual decomposition naturally invites the following intuition: if retrieval is perfect, a model should, in principle, perform just as well on a long input as it does on a short one.
Accordingly, failures in long-context tasks are often attributed to suboptimal or hallucinated retrieval. As a result, the ability to identify relevant evidence has been treated as a crucial capability for long-context LLMs and has shaped, at least in part, both the evaluation \interalia{kamradt2023needleinahaystack,Xiao2024RARbRA,mohtashami2023landmarkattentionrandomaccessinfinite,modarressi2025nolimalongcontextevaluationliteral,yu2025sequentialniahneedleinahaystackbenchmarkextracting} and model designs in recent efforts to improve long-context LLMs \interalia{Yu2023TrainingW,peng2023yarnefficientcontextwindow,li2023how,Xiong2024FromAN,Jin2024LongContextLM,fu2024dataengineeringscalinglanguage,ge2024littlegoeslongway,chen2024longloraefficientfinetuninglongcontext,xu2025128k4mefficienttraining,han2024lminfinitezeroshotextremelength}.

This work provides evidence that calls this premise into question.
Our systematic, controlled experiments across 5 open- and closed-source models on math, question answering, and code generation tasks show that even when a model can perfectly retrieve \emph{all} the evidence---in the strictest possible sense, reciting all tokens with 100\% exact match---its performance still degrades substantially as input length increases~(\S\ref{sec:measuring}).
For example, Llama-3.1-8B Instruct, with a claimed 128K context length, is able to retrieve all evidence with exact matches for 970 of 1000 MMLU problems ~\cite{hendrycks2021measuring} extended to 30k tokens with irrelevant tokens --- matching its retrieval performance on the same problems presented in the original form with shorter contexts. 
However, despite this robustness in retrieval, its accuracy drops by 24.2\% compared to the short-context case. 
More concerningly, this failure occurs even when the irrelevant tokens in the long context consist of minimally distracting whitespace (\cref{fig:failure-to-solve-problem}; \S\ref{subsec:space}), and even when the evidence is placed immediately before the question. Surprisingly, in a separate experiment, we observe a similar performance drop even when all irrelevant tokens are masked and the model attends \emph{only} to the evidence and the question---identical to those in the short-context setting except for the longer distance between the evidence and the question (\S\ref{subsec:space}).

These findings reveal a previously-unrealized limitation: the sheer length of the input alone can hurt LLM performance, independent of retrieval quality and \emph{without} any distraction.
They motivate the following hypothesis: even when retrieval is perfect, the model's performance can still be improved by limiting the number of tokens in the input. 
Our controlled experiments provide evidence supporting this hypothesis, and yield a simple and effective retrieve-then-reason mitigation strategy (\S\ref{sec:solution}). 
Specifically, we prompt the model to recite the evidence retrieved from the long context and prepend it directly before the question to form a new, shorter prompt to get the final output; this effectively converts the long-context task into a short-context one.
Our experiments on GPT-4o on RULER show that this simple approach consistently improves performance by up to 4\% on top of an already high baseline performance.

Our findings reveal previously underappreciated limitations in how current models approach long-context tasks.
They offer a potential explanation for a recurring observation in retrieval-augmented generation (RAG) that performance often saturates or even degrades as more documents are added to the context \interalia{Cuconasu2024ThePO,Jin2024LongContextLM, yu2024defenserageralongcontext},
and for recent findings that long CoTs can sometimes hurt the performance \citep{zeng2025revisitingtesttimescalingo1like}.
These results call for a rethinking of how long-context capabilities are evaluated.
In particular, benchmarks that isolate retrieval as a standalone capability might overestimate progress, as improvements in retrieval alone do not necessarily translate into better long-context performance; instead long-context capabilities should be evaluated holistically.
Practically, our proposed mitigation strategy is model-agnostic, simple, and effective.

\section{Background and Related Work}

Recent work on long-context LLMs has largely followed a dichotomy of long context capabilities: (1) retrieving relevant information from long inputs, and (2) solving the task using the retrieved evidence \interalia{qiu2025elicitingincontextretrievalreasoning,li2024alr2retrievethenreasonframeworklongcontext,zhang2025attentionrevealstokenstrainingfree,wu2024retrievalheadmechanisticallyexplains}.  This intuition motivates evaluation methods based on retrieval, such as needle-in-the-haystack tests \cite{kamradt2023needleinahaystack} and passkey retrieval \cite{mohtashami2023landmarkattentionrandomaccessinfinite}. The core intuition suggests that if a model can accurately retrieve the relevant formation, it should be able to use that information as effectively as it would in a short-context setting. From this perspective, improvements in retrieval are often taken as evidence of progress in long-context capabilities~\interalia{peng2023yarnefficientcontextwindow,Xiong2024FromAN,Jin2024LongContextLM,fu2024dataengineeringscalinglanguage,chen2024longloraefficientfinetuninglongcontext,xu2025128k4mefficienttraining,lin2025s2attentionhardwareawarecontextsharding}.

To better reflect real-world use cases, later benchmarks  extend this setup to include reasoning tasks that require aggregating multiple pieces of evidence, such as multi-step inference, variable binding, and multi-document question answering~\interalia{wang2025reasoningmultipleneedleshaystack, hsieh2024ruler, kuratov2024babilong, ling2025longreason, wang2024loong, li2024needlebenchllmsretrievalreasoning, song2024countingstarsmultievidencepositionawarescalable, zhang-etal-2024-bench}. Findings from these evaluations show that strong performance on synthetic retrieval tests does not always translate to more complex long-context tasks \interalia{hengle2025llmsreasonextendedmultilingual,Lee2024LongContext,an2024doeseffectivecontextlength}, suggesting a different conclusion that language models struggle to use information in long-context inputs as effectively as when the information is contained in a short-context \interalia{zhang2024middlelanguagemodelsuse, kuratov2024babilong, zhang2025leveraging}. These failures are typically attributed to suboptimal retrieval. For example, retrieval performance often drops when more distractors are present \cite{ivgi2022efficientlongtextunderstandingshorttext, goldman-etal-2024-really}, when retrieval and aggregation of multiple pieces of evidence is required~\cite{wang2025reasoningmultipleneedleshaystack, nocha-2024-karp-thai-et-al, agrawal2024evaluatingmultilinguallongcontextmodels, song2024countingstarsmultievidencepositionawarescalable}, when relevant passages have low lexical overlap \cite{modarressi2025nolima},  or when the evidence appears near the middle of the context \cite{liu-etal-2024-lost}. Some studies have also noted that irrelevant tokens can distract the model and impair its reasoning \cite{pmlr-v202-shi23a,wu2024easilyirrelevantinputsskew}. 

A deeper understanding of these failures---and LLMs long-context capabilities in general---requires carefully controlled experiments that disentangle factors such as context length, token-level distraction, and task complexity. To the best of our knowledge, such analysis remains scarce, yet it is essential for uncovering the true limits of current models and guiding future efforts.
This work takes steps toward that goal through a series of controlled experiments designed to address a central question: what prevents a model from solving a problem when it already has perfect access to all the information it needs? While the retrieval performance and the distractions from irrelevant content are both important, our findings reveal a previously overlooked factor: the sheer length of the context itself. 
Our findings complement the prevailing conclusions that long-context performance is often bottlenecked by retrieval failures or distraction. 
\section{Measuring Long-context Performance under Perfect Retrieval}
\label{sec:measuring}

To better understand the factors limiting LLMs' long-context performance,
\S\ref{subsec:benchmark} presents a series of systematic, controlled experiments designed to answer a simple but fundamental question:
\emph{When a model can perfectly retrieve all the information it needs, can it solve long-context tasks as effectively as short-context ones?}
We observe a consistent and substantial performance drop across 5 open and closed models even when all evidence can be retrieved with a 100\% exact match (\S\ref{subsec:performance}).

\subsection{A Long-Context Synthetic Benchmark Covering Math, QA, and Coding}\label{subsec:benchmark}

This section introduces our experiment setting and lays the ground work for onward discussion. Our synthetic benchmark is constructed by forming long-context tasks from short-context ones, inspired by recent efforts \cite{Bai2024LongAlignAR,Wu2024LongCA,Hu2024LongRecipeRF,Zhu2025GeneralizingFS}.
We include math, question answering, and code generation tasks to make our conclusions more relevant to a broad range of application scenarios.

\begin{figure*}[htbp]
    \centering
    \includegraphics[width=1.0\linewidth]{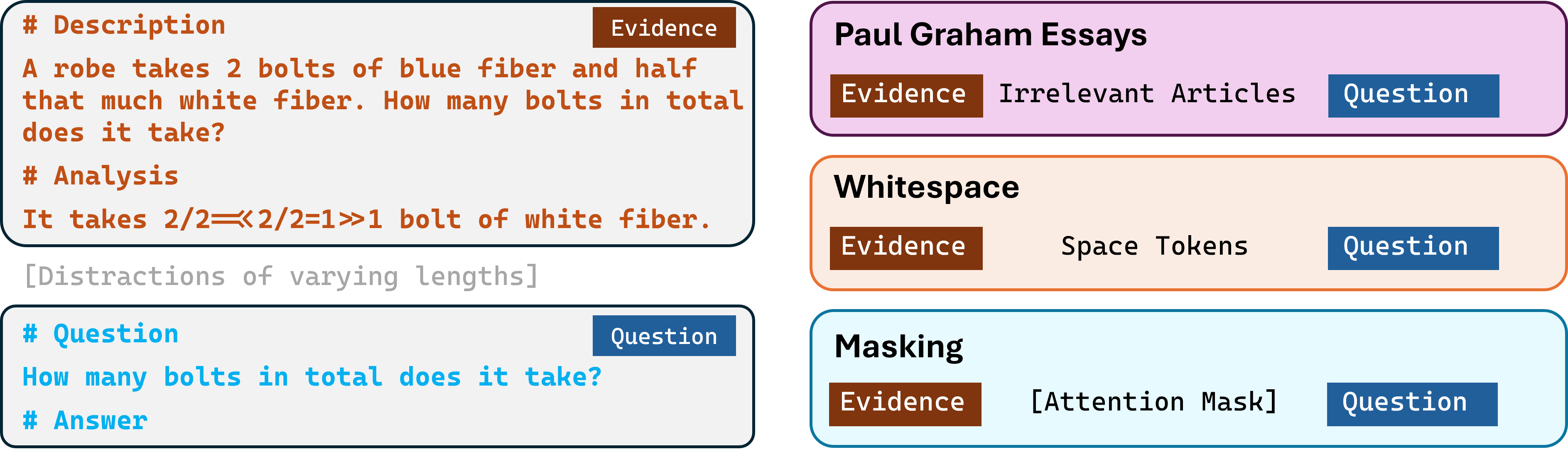}
    \caption{\textit{Left}: In our synthetic benchmark, each long-context problem is created by separating a short-context problem into evidence and question, and extending the length with distraction tokens. \textit{Right}: We discuss three types of distractions in this work, ordered by decreasing strength: Essay tokens (\cref{sec:measuring}), Whitespace (\cref{subsec:space}), and masking out all distraction tokens (\cref{subsec:masking}).}
    \label{fig:filler_experiment}
\end{figure*}

\Cref{fig:filler_experiment} provides an illustrative example. 
We identify two components of each problem: the \term{evidence} and the \term{question}. The evidence contains \emph{all} the information that the model needs to solve the task, and the question contains the query and format requirements. 
Both are specific to the task and will be detailed later in this section. 
Given a pair of evidence and question, 
we insert \term{distraction tokens} in between to reach desired context lengths. 
This creates input of the form: 
\texttt{[Evidence] [Distraction Tokens] [Question]}. 
Intuitively, this setting simulates real-world scenarios where a user interacts with a chatbot over a long dialogue, and the model must retrieve evidence from an earlier part of the conversation to answer the current query.

While previous benchmarks increase the difficulty of retrieval through different methods such as scattering the evidence across the context \cite{kuratov2024babilong, hsieh2024ruler}, and placing the evidence in different positions \cite{kuratov2024babilong, hsieh2024ruler, zhang-etal-2024-bench, wang2024loong}, 
we aim to make retrieval as easy as possible to control for perfect retrieval in order to answer our RQ.
This leads to two of our major design choices: 
(1) We keep the evidence in a single consecutive chunk to avoid requiring the model to aggregate scattered evidence.
(2) We intentionally place the evidence at the beginning of the input, which, as the Lost-in-a-Middle effect~\cite{liu-etal-2024-lost} suggest, is the easiest location to retrieve; the question is put at the end to better simulate real-world applications.
In this experiment we follow \citet{kamradt2023needleinahaystack} and use 
Paul Graham Essays \cite{paulgrahamessays} as the distraction tokens.
Our setting generalizes to different evidence locations and distraction tokens, which we explore later in \S\ref{sec:sheer_length}.

\paragraph{Diverse tasks covering math, QA, and coding}

To isolate the effect of context length on model performance, we intentionally select tasks that are commonly used to evaluate LLMs and on which most models perform consistently well, at least under short-context settings.
To cover different capabilities, we use the following datasets: math~(GSM8K;~\citealp{cobbe2021trainingverifierssolvemath}), question answering~(MMLU;~\citealp{hendrycks2021measuringmassivemultitasklanguage}), and code generation~(HumanEval;~\citealp{chen2021codex}). 
In addition, we include a synthetic Variable Summation task in order to show whether this degradation applies to very simple problems. This task is inspired by Variable Tracking \cite{hsieh2024ruler}, Distractor Variable Assignment \cite{li2025questbenchllmsaskright} and RV-Bench \cite{hong2025benchmarkinglargelanguagemodels}, and aims to test the model’s ability to perform very basic arithmetic operationss: summing a subset of variables from a given list.
\cref{tab:task_summary} summarizes the evidence and question of these tasks.

\begin{table}
\centering
\small 

\renewcommand\cellalign{cc}
\newcolumntype{C}[1]{>{\centering\arraybackslash}m{#1}}
\begin{tabular}{@{} C{1.5cm} 
                C{1.2cm} 
                C{2cm} 
                C{1.5cm} 
                @{}}
\toprule
\textbf{Task Name} & \textbf{Type} & \textbf{Evidence} & \textbf{Question}\\
\midrule
Variable Summation (VarSum) & Variable Tracking & Values of 50 integer variables & The sum of 3 random variables \\\midrule
GSM8K & Math & Problem description with chain-of-thought steps & Specific question \\\midrule
MMLU & Multiple Choice QA & Problem description & Question and four options \\\midrule
HumanEval & Coding & Function definition w/ docstring & Instruction  \\
\bottomrule
\end{tabular}
\caption{Summary of our tasks targeting different types of model capacities, with the identification of evidence and question. The test sets of GSM8K, MMLU and HumanEval are used.
}
\label{tab:task_summary}
\end{table}

\paragraph{Measuring retrieval with exact match}

To quantify retrieval performance, we prompt the model to recite both the evidence and the question exactly as they appear in the input. Retrieval performance is measured using \term{exact match}, where the model receives a score of zero if there is a difference between its output and the original evidence or question. Note that the retrieval is evaluated separately from the actual performance evaluation. 
In the latter set-up, we do not ask the model to recite before solving the problem.

In this work, we intentionally measure retrieval in the strictest possible way to rule out the effect of retrieval failures as a source of error in problem solving, in contrast to existing studies~\interalia{li2024alr2retrievethenreasonframeworklongcontext, kuratov2024babilong, li2024needlebenchllmsretrievalreasoning, qiu2025elicitingincontextretrievalreasoning}. Although these works report a similar gap between long-context retrieval and reasoning performance, some  
compare retrieval and problem-solving on different tasks; in such settings, a perfect score on a retrieval task does not ensure that there are no retrieval failures on the actual problem-solving tasks. Others, which involve in their settings retrieval, aggregation and relatively simple reasoning over multiple evidence pieces (needles), suggest multiple causes for the gap, such as the model's failure to extract or aggregate all needles. In both types of studies, retrieval cannot be conclusively excluded as a possible failure mode.

In contrast, by measuring retrieval on the evidence for the exact problems they are tasked to solve, we are allowed to isolate the effect of retrieval failures and conduct a more direct investigation of our research question: the effect of the sheer context length on problem solving. To the best of our knowledge, our work is among the first attempts to measure long-context performance under explicit control for perfect retrieval.

\subsection{Performance Drops Despite Perfect Retrieval}
\label{subsec:performance}

\paragraph{Models}
We choose two open-source models, Llama-v3.1-8B-Instruct and Mistral-v0.3-7B-Instruct, because of their long-context capability (with 128K and 32K claimed context lengths respectively) and good performance on short-context benchmarks \cite{grattafiori2024llama3herdmodels, jiang2023mistral7b}. 
Both models are widely used for post-training, while having different architectures. This choice of the two representative but very different models can help make
our conclusions more practically relevant in broader applications.

\paragraph{Results}

\begin{figure*}[htbp]
    \centering
    \includegraphics[width=0.6\linewidth]{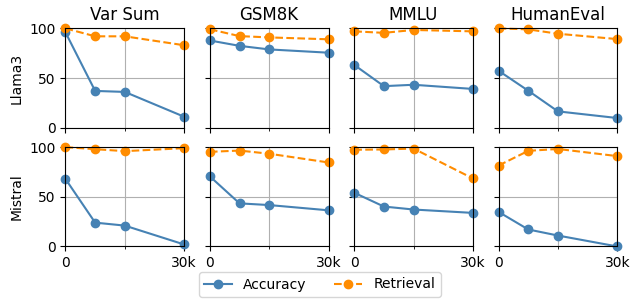}
    \caption{Evaluation results on Llama3-8B and Mistral-v0.3-7B, with performance accuracy in problem solving
(Accuracy) and retrieval scores measured by Exact Match (Retrieval). "Context length" refers to the \textit{total number of input tokens} for each problem, which is crafted by inserting PaulGrahamEssay tokens between \texttt{evidence} and \texttt{question} (as illustrated in \cref{fig:filler_experiment}). See Appendix for detailed numbers.}
    \label{fig:essay_retrieval_results}
\end{figure*}

\cref{fig:essay_retrieval_results} shows the results of the retrieval and problem solving performance. Although there exists a drop in retrieval scores, this drop is relatively marginal until the length reaches 30K tokens. In fact, for inputs shorter than 15K tokens, both models are able to accurately extract the problem description except for no more than 8.2\% of the problems.  Note again that the retrieval score is calculated by exact match and a failure does not necessarily mean the model cannot extract the evidence. 

In contrast, accuracy drops drastically across all tasks, by a larger margin as opposed to the retrieval score across almost all tasks and context lengths (except Llama3 on GSM8K with 7k tokens). A large portion of the drop in problem solving happens within 7k tokens, well below the limit of either model where retrieval performance starts to degrade. On Var Sum, for example, the number is 59\% off the baseline 96\% for Llama, and 44\% off the 0-context 68\% for Mistral, while the retrieval scores only drop by 8\% and 2\% respectively; on HumanEval, the retrieval scores for Mistral even increase on longer inputs while its accuracy scores keep decreasing.  

\paragraph{Discussion}
Our results confirm that even in the cases where a model accurately retrieves all the evidence, it may still fail to solve a long-context task of which it is capable of solving the short-context version. This  align with existing ones in reporting a performance drop under long input \cite{li2024alr2retrievethenreasonframeworklongcontext, kuratov2024babilong} while providing fresh insights.
Existing conclusions often entangle retrieval accuracy with task performance. 
Common settings like locating relevant info from distracting text and reasoning through aggregated evidence may fail due to retrieval, aggregation (not discussed here), reasoning, or the input length itself.
This work, to the best of our knowledge, for the first time, presents systematic evidence suggesting that the model's capabilities in reasoning, QA, and coding degrade with longer inputs and contribute to their failures, even when retrieval is perfect.
Our findings \emph{never} seek to diminish the importance of the retrieval;
rather, by simulating an ``upper bound'' --- perfect retrieval --- they raise a complementary question that is often overlooked:
in addition to improving models' ability to retrieve the right information, we must also ask \emph{can the model still use that information effectively in long-context settings?}

\begin{table}[ht]
\centering
\scriptsize
\begin{tabular}{@{} llccccc @{}}
\toprule
Model & Task & 0 & 7500 & 15000  & 30000 \\
\midrule
\multirow{4}{*}{GPT-4o}
  & VarSum     & 100.0 & 0.0  & 0.0   & 0.0 \\
  & GSM8K      & 87.8 & -7.0 & -8.5  & -7.0 \\
  & MMLU       & 82.4 & -2.1 & -0.3 & -1.0 \\
  & HumanEval  & 68.3 & 0.0  & 0.0   & -3.1 \\
\midrule
\multirow{3}{*}{Claude-3.5}
  & VarSum     & 90.2 & -0.6 & -5.4  & -4.8 \\
  & GSM8K      & 95.3 & -3.8 & -5.2  & -6.0 \\
  & MMLU       & 82.2 & -41.7 & -38.8 & -67.6 \\
  & HumanEval  & 90.2 & -0.6 & -5.4  & -4.8 \\
\midrule
\multirow{4}{*}{Gemini}
  & VarSum     & 100.0 & 0.0  & 0.0    & 0.0 \\
  & GSM8K      & 83.2 & +7.7 & +8.6  & +6.2 \\
  & MMLU       & 81.9 & -3.0 & -3.5  & -3.9 \\
  & HumanEval  & 86.0 & -11.0 & -2.5  & -1.8 \\
\bottomrule
\end{tabular}
\caption{Performance drop across different lengths on selected closed-source models, with corresponding numbers of whitespace tokens \textit{between} \texttt{evidence} and \texttt{question}. }
\label{tab:closeapi-delta}
\end{table}

\begin{figure*}[htbp]
    \centering
    \includegraphics[width=0.7\linewidth]{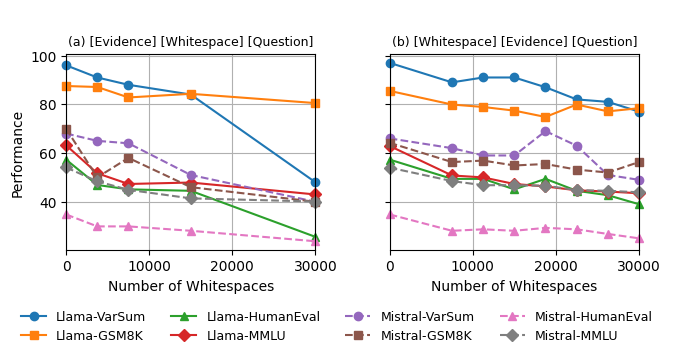}
    \caption{
        Performance across different context lengths on Llama-3-8B Instruct and Mistral-v0.3-7B-Instruct, with corresponding numbers of whitespace tokens inserted for minimum distraction. (a, Left) Whitespaces are inserted \textit{between} \texttt{evidence} and \texttt{question}.  (b, Right) Whitespaces are inserted \textit{before} \texttt{evidence}, and \texttt{question} adjacent to \texttt{evidence}.
    }
    \label{fig:space_llama_mistral}
\end{figure*}

\section{Models Struggle Even Without Distraction}\label{sec:sheer_length}
This section aims to answer 
\emph{What prevents a model from effectively using information it has successfully retrieved?}
In addition to the distraction from the irrelevant tokens, which aligns with existing observations \cite{pmlr-v202-shi23a,wu2024easilyirrelevantinputsskew},
we reveal a surprising finding: the sheer length of the context alone can negatively impact the model's performance, even when there is little (\S\ref{subsec:space}) to no (\S\ref{subsec:masking}) distraction.

\subsection{Performance Degradation with Minimum Distraction}\label{subsec:space}

To reduce the distraction from irrelevant tokens, we modify the benchmark design in \S\ref{subsec:benchmark} , by replacing the natural language tokens with whitespace; 
all other settings are kept the same. 
We intuitively choose whitespace, since it generally carries minimum information and is a natural separator, creating least distraction \citep{zhang2025understandingrelationshippromptsresponse}.

\paragraph{Llama and Mistral}
\cref{fig:space_llama_mistral}(a) shows our results for Llama and Mistral. Although these results generally reflect an improvement compared to those under the essay distraction in the previous section (\cref{fig:essay_retrieval_results}), we can still see a substantial drop in performance for both models and all tasks: at least 7\% at 30K space tokens (as in Llama-GSM8K), and more significant drops, notably including a 48\% drop for Llama on VarSum and 30\% for Mistral on GSM8K. 

\paragraph{Closed-source Models}
We also test three closed-source models: GPT4o \cite{openai2024gpt4technicalreport}
, Claude-3.7-Sonnet\cite{anthropic2024claude3} and Gemini-2.0 \cite{google2024gemini2} on our selected tasks. 
Results are shown in \cref{tab:closeapi-delta}. We observe a very different pattern with these models from their smaller open-source counterparts. They experience a smaller drop across increased context lengths. For VarSum, both GPT-4o and Gemini-2.0 achieve perfect performance throughout. The closed-source models generally exhibit more robustness than the open-source ones in terms of the negative impact of context lengths. However, substantial and mostly consistent degradation is still observed in most models and tasks, despite varying trends among tasks - with the notable exception of Gemini on GSM8K, where performance at 30K actually improves by 8.6\%, and HumanEval, where the performance improves after a certain length in some cases (15K vs 7K for Gemini, for example). 

To determine if context length itself is the factor, we also need to control the relative distance between evidence and question, as existing works \cite{li2024alr2retrievethenreasonframeworklongcontext, an2024doeseffectivecontextlength} suggest that it may affect performance drop. Therefore, we move the evidence back to the end of the input, right before the question, so that the distance does not change with input size. Our results are shown in \cref{fig:space_llama_mistral}(b), where a substantial drop is observed despite occasional fluctuations: up to 17\% for Mistral and 20\% for Llama under 30K space tokens.

Previous observations like Lost-in-the-middle \cite{liu-etal-2024-lost} acknowledge the affect of evidence position in long-context performance, especially when the evidence is in the middle of the text; on the other hand, our results prove that the performance degradation is directly  related to the input length alone, regardless of the relative position between evidence and question. In fact, the degradation still happens when the evidence is put in the best positions possible, the beginning and the end of the text, and that further strengthens that the sheer length of input is a decisive factor to the degradation.

\subsection{Eliminating Distraction Completely with Masking}\label{subsec:masking}
The previous experiment with whitespace already provides initial implication that with minimal distraction, the models are still hurt by increased context size. Now, we take one step further and seek \emph{no} distraction, by masking \emph{all} distraction tokens when calculating attention for our targeted open-source models, Llama and Mistral.
Effectively, the input to the model becomes \texttt{[Evidence] [Masks] [Question]}, where the model attends \emph{only} to the evidence and the question, identical to the short-context setting except for the increased distance between them introduced by the masked tokens.
 The results are in \cref{tab:masking-filler}. Surprisingly, yet still in tune with our expectations, we still observe a consistent performance drop, which reaches at least 7.9\% for both models at 30K masked distraction tokens. Some drops are even larger compared to when we fill the context with space: for HumanEval, Llama3 suffers a 50\% drop with masking compared to that of only 19.4\% with space.
 
\begin{table}[ht]
\centering
\scriptsize

\begin{tabular}{@{} llrrrrr @{}}
\toprule
Model & Task & 0 & 3750 & 7500 & 15000 & 30000 \\
\midrule
\multirow{4}{*}{Llama3}
 & VarSum    & 97.0  & -11.0  & -35.0  & -24.0  & -50.0 \\
 & GSM8K     & 86.1  &  -1.7  &  -3.3  &  -4.3  & -19.6 \\
 & MMLU      & 62.8  & -11.3  & -15.9  & -15.5  & -21.1 \\
 & HumanEval & 57.3  &  -5.5  & -22.0  & -16.5  & -50.0 \\
\midrule
\multirow{4}{*}{Mistral}
 & VarSum    & 66.0  &  -5.0  & -11.0  & -19.0  & -34.0 \\
 & GSM8K     & 64.5  &  -2.1  &  -4.8  &  -8.2  & -15.1 \\
 & MMLU      & 53.8  &  -4.7  &  -7.5  & -11.0  & -11.8 \\
 & HumanEval & 34.8  &  -7.3  &  -8.5  & -10.4  &  -7.9 \\
\bottomrule
\end{tabular}
\caption{Llama-3 and Mistral still suffer a performance drop with increased length even when all distractions are masked. The numbers 0, 3750, etc. are lengths of masked distraction in tokens.  }
\label{tab:masking-filler}
\end{table}

\paragraph{Discussion}

Through these settings, we feel more confident to conclude that long-context language models suffer a common performance degradation when solving long-context tasks, even with perfect retrieval, even with minimum or zero distraction. 
Our conclusion suggests limitations for practical applications. For example, in typical scenarios like chatbot dialogues, even when the question immediately follows its evidence or evidence is pinpointed, longer input may still lead to unexpected failures.
Our finding also provides insight for the actual mitigation of the long-context degradation. One incentive that naturally emerges is to simply shorten the length of the input context. In the next section, we shall present a proof-of-concept solution based on this idea.

\section{Shortening Input Through Retrieval: A Simple Fix}\label{sec:solution}

Having learned the negative impact of the sheer length of the context (\S\ref{sec:sheer_length}), we naturally arrive at the following hypothesis:
\emph{Even when retrieval is perfect, a model’s performance can still be improved by limiting the number of tokens in the input.}

This section introduces a simple, model-agnostic, and effective mitigation strategy for cases where accurate retrieval---though not necessarily perfect---can be achieved. We then present experimental results that support the above hypothesis.

\begin{figure}[htbp]
    \centering
    \includegraphics[width=0.9\linewidth]{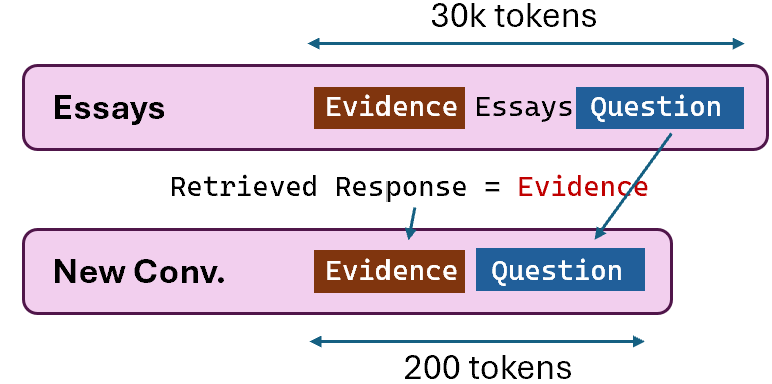}
    \caption{Our strategy retrieves \texttt{evidence} to shorten the context length before solving the task.}
    \label{fig:retrieve_then_reason}
\end{figure}

\paragraph{Retrieve then Solve }
Given a long-context input problem, 
our strategy first prompts the model to retrieve and recite all relevant information from the input context. This recited evidence is then concatenated with the original problem statement to form a new, shorter prompt. 
The model solves the problem based solely on the recited evidence, without having the long context as part of its input, similarly to starting a new chat session in ChatGPT (see \cref{fig:retrieve_then_reason} for an illustrative example).
This effectively turns a long-context task into a short-context one using an additional prompt.

This approach is related to \citet{li2024alr2retrievethenreasonframeworklongcontext}, which improves the model by training it to align to both retrieval and reasoning objectives. Our method, in contrast, does not address the retrieval problem itself; rather, with an explicit retrieval step, it assumes accurate retrieval as a prerequisite. 

\paragraph{Experiments}

We evaluate our strategy on two benchmarks: (1) our synthetic benchmark under the initial setting, where we insert Essay tokens between evidence and question of GSM8K problems. (2) Two QA tasks, QA1 and QA2, of RULER \cite{hsieh2024ruler}, an established long-context benchmark, in which models are provided with a problem and a number of potentially related documents, and are required to retrieve the answer from one or more of the documents. We compare our strategy against the baseline method where the model is directly asked to answer the question based on the input.

On our synthetic benchmark, as shown in \cref{tab:reasoning_context}, with our method, Mistral-v0.3-7B Instruct achieves a substantial performance boost under longer contexts without excessive prompt engineering, with a gap of less than 10\% and a 30\% boost until the input size reaches 26K.
    
On QA1 and QA2 tasks of RULER, we experiment with GPT-4o, taking advantage of its retrieval capabilities. 
As shown in \cref{tab:qa-accuracy}, under varying context lengths ranging from 128K to 4K tokens, while the baseline already achieves strong performance on QA1 (88.2–90.4\%), our method yields consistent improvements, reaching 92.2\% at 4K.  
 Our method achieves a larger improvement on QA2 by maximum 4\% at 32K.

\begin{table}[ht]
\centering
\small
\begin{tabular}{crrrrr}
\toprule
\textbf{Length} & 0 & 3750 & 7500 &  15000  & 26250  \\
\midrule
Baseline & 70.6 & 49.3 & 43.4 &   41.6 &  35.5  \\
Ours   & 76.2 & 71.4 & 66.7 &  69.1 & 66.7  \\
\bottomrule
\end{tabular}
\caption{On our synthetic benchmark with essay distractions, we improve performance of Mistral-v0.3-7B Instruct by up to 31.2\% on GSM8K. }
\label{tab:reasoning_context}
\end{table}

Our results suggests that this simple and model-agnostic approach can enhance a model’s ability to make use of the information they can accurately retrieve from long contexts. In doing so, it helps close the gap between improvements in retrieval performance and actual gains on long-context tasks.

\begin{table}[ht]
\centering
\scriptsize
\label{tab:qa-accuracy}
\begin{tabular}{@{} llcccccc @{}}
\toprule
Method & Task & 128K & 64K & 32K & 16K & 8K & 4K \\
\midrule
\multirow{2}{*}{Baseline}
  & QA1 & 88.2 & 87.8 & 87.8 & 88.8 & 87.2 & 90.4 \\
  & QA2 & 63.2 & 67.0 & 68.4 & 69.4 & 71.4 & 71.2 \\
\midrule
\multirow{2}{*}{Ours}
  & QA1 & 88.2 & 88.4 & 88.6 & 89.8 & 89.8 & 92.2 \\
  & QA2 & 65.4 & 70.6 & 72.4 & 72.8 & 74.0 & 73.2 \\
\bottomrule
\end{tabular}
\caption{Our method consistently improves the performance of GPT-4o on tasks QA1 and QA2 of the RULER benchmark.}
\end{table}

\section{Discussion}

Our observations on the performance drop, even with masking, along with the relatively steady performance of closed-source models, are related to findings from \cite{li2024alr2retrievethenreasonframeworklongcontext, an2024doeseffectivecontextlength} 
which attribute the drop to a distribution bias with position introduced during training. While \citet{an2024doeseffectivecontextlength} targets the drop in retrieval and \citet{li2024alr2retrievethenreasonframeworklongcontext} addresses the long-context performance as a whole, our work further shows that this cause also applies to the degradation caused by the length itself, regardless of retrieval or distraction strength.  

Our results further imply that the previous two-part decomposition of long-context problem solving into retrieval and problem solving  \cite{qiu2025elicitingincontextretrievalreasoning,li2024alr2retrievethenreasonframeworklongcontext,zhang2025attentionrevealstokenstrainingfree} is inconclusive, urging researchers to further explore the underlying mechanisms. 
The current effort to independently focus on improving models' long-context retrieval and short-context capacities at training \cite{grattafiori2024llama3herdmodels,megabeam-mistral-7B-512k-2024,yang2025qwen3technicalreport,ai2024yi} may not fully translate to an overall improvement in long-context ability: failure is possible despite the model excelling in both.
Our conclusion encourages a more comprehensive evaluation beyond focusing on retrieval, with a more fine-grained analysis to precisely separate each failure mode. 

Our conclusion supplements existing observations for practical applications. It supports previous findings (e.g., \citealt{li2024longcontextvsrag, yu2024defenserageralongcontext}) that RAG suffers from retrieving too many documents. It also aligns with \citealt{dai2025sgrpoearlyexitreinforcement, zeng2025revisitingtesttimescalingo1like}, which argue that generating excessively long CoTs can hurt reasoning models, despite being a common strategy.

\section{Conclusion}

In this work, we expose a previously noticed but unexplored limitation: the performance degradation of language models may be attributed to the length of the input itself, even when the model is able to retrieve all relevant information, and all distractions are removed. Our findings challenge the popular view of decomposing long-context task solving into retrieval and problem solving, and encourage more consideration on future model designs and evaluations. Our simple yet effective strategy shows that the degradation can be mitigated through reducing the context length, serving as an initial attempt in bridging the gap between retrieval and long-context performance.
\section*{Limitations}

The conclusions of this work are only based on two open-source models, three closed-source models and 4 tasks, despite consciously selecting those that are more representative. We did not run experiments for some combinations of settings, such as retrieval on closed-source models (due to these models occasionally refusing to recite evidence under a long input). 

The method proposed in Section 5 has limited use case. It requires perfect retrieval, which is already hard in many real-world tasks that feature retrieval settings harder than in our synthetic data. We did not report results for open-source models on RULER, due to the fact that the failure of retrieval directly causes a performance drop (compared to baseline), and it is not our focus in this paper to address this type of failure.

\section*{Declaration of generative AI and AI-assisted technologies}

While preparing this work, we used OpenAI's Codex, GPT-4o, o3 and ChatGPT 5 to assist with coding, as well as perform spell/grammar check, word selection, and rephrasing in paper writing. We inspected the generated contents and edited them as needed. We take full responsibility for the content of this work.

\section*{Acknowledgments}

This research was supported by the Amazon-Illinois Center on AI for Interactive Conversational Experiences (AICE Center). 
E.A.H. acknowledges support from NSF grants OAC-2514142 and OAC-2209892. This work was supported by Laboratory Directed Research and Development (LDRD) funding from
Argonne National Laboratory, provided by the Director, Office of Science, of the U.S. Department
of Energy under Contract No. DE-AC02-06CH11357. An award for computer time  was provided by the U.S. Department of Energy’s
Innovative and Novel Computational Impact on Theory and
Experiment (INCITE) Program. This research used supporting
resources at the Argonne and the Oak Ridge Leadership
Computing Facilities. The Argonne Leadership Computing
Facility at Argonne National Laboratory is supported
by the Office of Science of the U.S. DOE under Contract
No. DE-AC02-06CH11357. The Oak Ridge Leadership
Computing Facility at the Oak Ridge National Laboratory
is supported by the Office of Science of the U.S. DOE
under Contract No. DE-AC05-00OR22725. This research used
both the DeltaAI advanced computing and data resource,
which is supported by the National Science Foundation
(award OAC 2320345) and the State of Illinois, and the
Delta advanced computing and data resource which is
supported by the National Science Foundation
(award OAC 2005572) and the State of Illinois.
Delta and DeltaAI are joint efforts of the
University of Illinois Urbana-Champaign and
its National Center for Supercomputing Applications.

\bibliography{custom}
\appendix
\section{Appendix} 

\subsection{System Prompt}

See Figures \ref{fig:prompt_gsm8k}, \ref{fig:prompt_gsm8k_ret}, \ref{fig:prompt_varsum}, \ref{fig:prompt_varsum_ret}, \ref{fig:prompt_mmlu}, \ref{fig:prompt_mmlu_ret}, \ref{fig:prompt_ruler}, \ref{fig:prompt_ruler_example}.

\begin{figure}[hptb]
\centering
\noindent\fbox{
    \parbox{\linewidth}{
    \# Problem Description \\
     \{problem description\}\\
    \# Analysis\\
    \{cot\}\\
    \# Others\\
    \{distraction\}\\
    \# Question\\
    \{question\}\\
    \# Answer\\
    }
}
\caption{Prompt for GSM8K Problems.}
\label{fig:prompt_gsm8k}
\end{figure}

\begin{figure}[htbp]
\centering
\noindent\fbox{
    \parbox{\linewidth}{
    \# Problem Description \\
    \{problem description\} \\
    \# Analysis \\
    \{cot\} \\
    \# Others \\
    \{distraction\} \\
    \# Question \\
    \{question\} \\
    \# Answer \\
    Let's first recite ``\# Problem Description'', ``\# Analysis'' and ``\# Question'' word by word, and then think and answer in the ``\#\# Answer'' subsection. Your response will be compared to the original question using exact match. When reciting, do not alter the original text.\\
    \#\# Problem Description
    }

}
    \caption{Prompt for GSM8K Problems, Retrieval Task.}
\label{fig:prompt_gsm8k_ret}
\end{figure}

\begin{figure}[htbp]
\centering
\noindent\fbox{
    \parbox{\linewidth}{
    \# Problem Description \\
    \{evidence\} \\
    \# Others \\
    \{distraction\} \\
    \# Question \\
    \{question\} \\
    \# Answer \\
    Let's think step by step.
    }
}
    \caption{Prompt for VarSum Problems.}
\label{fig:prompt_varsum}
\end{figure}

\begin{figure}[htbp]
\centering
\noindent\fbox{
    \parbox{\linewidth}{ 
    \# Problem Description \\
    \{evidence\} \\
    \# Others \\
    \{distraction\} \\
    \# Question \\
    \{question\} \\
    \# Answer \\
    Let's first recite ``\# Problem Description'' and ``\# Question'' word by word, and then think and answer in the ``\#\# Answer'' subsection. Your response will be compared to the original question using exact match. When reciting, do not alter the original text.\\
    \#\# Problem Description
    }
}
    \caption{Prompt for VarSum Problems, Retrieval Task.}
\label{fig:prompt_varsum_ret}
\end{figure}

\begin{figure*}[htp]
\centering
\noindent\fbox{
    \parbox{\linewidth}{ 
    \# Problem Description \\
    \{evidence\} \\
    \# Others \\
    \{distraction\} \\
    \# Question \\
    Choose the option that best satisfies the problem description.\\
    \{options\} \\
    Give only the number for the correct option. \\
    \# Answer \\
    The correct option is
    }
}
    \caption{Prompt for MMLU Problems.}
\label{fig:prompt_mmlu}
\end{figure*}

\begin{figure*}[htp]
\centering
\noindent\fbox{
    \parbox{\linewidth}{ 
    \# Problem Description \\
    \{evidence\} \\
    \# Others \\
    \{distraction\} \\
    \# Question \\
    Choose the option that best satisfies the problem description.\\
    \{options\} \\
    \# Answer \\
    Let's first recite ``\# Problem Description'' and ``\# Question'' word by word, and then think and answer in the ``\#\# Answer'' subsection.\\
    \#\# Problem Description
    }
}
    \caption{Prompt for MMLU Problems, Retrieval Task.}
\label{fig:prompt_mmlu_ret}
\end{figure*}

\begin{figure*}[htp]
\centering
\noindent\fbox{
    \parbox{\linewidth}{
    Read the question based on the given documents. Choose the documents most relevant to the question and copy them fully WORD BY WORD.\\
    \\
    The following are given documents.\\
    \\
    \{context\}\\
    Read the question. Choose the documents most relevant to the question and copy them fully WORD BY WORD. Only give me the relevant documents and do not output any other words. Do not make up documents. Do not repeat documents. Use ``\# Document <number>'' at the start of each document, then copy its full content.\\
    \\
    Question: \\
    \{question\}
    }
}
    \caption{Prompt for the first step of the Retrieval before Reasoning Strategy on RULER's QA1 and QA2 tasks. The second step uses the original prompt from RULER.}
\label{fig:prompt_ruler}
\end{figure*}

\begin{figure*}[htp]
\centering
\noindent\fbox{
\parbox{\textwidth}{ 

\texttt{<INPUT>}\\
(Berlin) Film Festival; and Best International Documentary at the Harlem (New York) International Film Festival".\\
Read the question. Choose the documents most relevant to the question and copy them fully WORD BY WORD. Only give me the relevant documents and do not output any other words. Do not make up documents. Do not repeat documents. Use ``\# Document <number>'' at the start of each document, then copy its full content.\\
\textbf{Question:} What type of film are both ``500 Years Later'' and ``Manson''?\\

\vspace{0.5em}
\texttt{<FIRST RESPONSE>}\\
\# Document 2\\
Manson (film)\\
Manson is a documentary film made in 1973 about Charles Manson and his followers. It was directed by Robert Hendrickson and Laurence Merrick.\\

\# Document 20\\
500 Years Later\\
500 Years Later ( ``500 amätat bähwala '') is an independent documentary film directed by Owen 'Alik Shahadah, written by M. K. Asante, Jr. and released in 2005. It has won five international film festival awards in the category of Best Documentary (including) UNESCO ``Breaking the Chains Award''). Other awards it has won include ``Best Documentary at the Pan African (Los Angeles) and Bridgetown (Barbados) Film Festivals; Best Film at the International Black Cinema (Berlin) Film Festival; and Best International Documentary at the Harlem (New York) International Film Festival''.\\

\vspace{0.5em}
\texttt{<SECOND INPUT>}\\
Answer the question based on the given documents. Only give me the answer and do not output any other words.\\

The following are given documents.\\
\# Document 2\\
Manson (film)\\
Manson is a documentary film made in 1973 about Charles Manson and his followers. It was directed by Robert Hendrickson and Laurence Merrick.\\

\# Document 20\\
500 Years Later\\
500 Years Later (``500 amätat bähwala '') is an independent documentary film directed by Owen 'Alik Shahadah, written by M. K. Asante, Jr. and released in 2005. It has won five international film festival awards in the category of Best Documentary (including) UNESCO ``Breaking the Chains Award''). Other awards it has won include ``Best Documentary at the Pan African (Los Angeles) and Bridgetown (Barbados) Film Festivals; Best Film at the International Black Cinema (Berlin) Film Festival; and Best International Documentary at the Harlem (New York) International Film Festival''.\\

Answer the question based on the given documents. Only give me the answer and do not output any other words.\\
\textbf{Question:} What type of film are both ``500 Years Later'' and ``Manson''?\\

\vspace{0.5em}
\texttt{<SECOND RESPONSE>}\\
Documentary film
}
}
\caption{An Example of the Retrieval before Reasoning Strategy on RULER.}
\label{fig:prompt_ruler_example}
\end{figure*}

\subsection{Detailed Results}

See Tables \ref{tab:acc&retrieval},  \ref{tab:space-filler}, \ref{tab:whitespace}, and \ref{tab:masking-filler-acc-ret}.

\begin{table*}[htp]
\centering  
\newcolumntype{C}[1]{>{\centering\arraybackslash}m{#1}}
\begin{tabular}{@{}C{2cm}C{2cm}C{1.5cm}rrrr@{}}
\toprule
\multirow{2}{*}{Model} & \multirow{2}{*}{Task} & \multirow{2}{*}{Metric} &
\multicolumn{4}{c}{Context Length (tokens)} \\ \cmidrule(lr){4-7}
 & & & 0  & 7500 & 15000 & 30000 \\
\midrule
\multirow{6}{*}{Llama3}
  & Var Sum & Acc.            &  96.0  & -59.0 & -60.0 & -85.0 \\
  &        &  Ret.  & 100.0  &  -8.0 &  -8.0 & -17.0 \\ 
  \cmidrule{2-7}
  & GSM8K  & Acc.            &  87.8  &  -5.4 &  -9.0 & -12.3 \\
  &        &  Ret.  &  99.1  &  -6.9 &  -8.2 & -10.1 \\
  \cmidrule{2-7}
  & MMLU   & Acc.            &  63.2  & -21.4 & -20.0 & -24.2 \\
  &        &  Ret.  &  97.0  &  -1.5 &   +1.4 &   0.0 \\
  \cmidrule{2-7}
  & HumanEval & Acc         &  57.3  & -20.1 & -40.9 & -47.6 \\
  &           & Ret.        & 100.0 & -1.0 & -5.4 & -10.8 \\
\midrule
\multirow{6}{*}{Mistral}
  & Var Sum & Acc.            &  68.0  & -44.0 & -47.0 & -66.0 \\
  &        &  Ret.  & 100.0 &    -2.0 &  -4.0 &  -1.0 \\\cmidrule{2-7}
  & GSM8K  & Acc            &  70.6  & -27.2 & -28.9 & -34.2 \\
  &        &  Ret.  &  95.3 &     +1.3 &  -1.9 & -10.7 \\\cmidrule{2-7}
  & MMLU   & Acc            &  54.1  & -13.9 & -16.9 & -20.3 \\
  &        &  Ret.  &  97.4 &   +0.3 &   +1.2 & -28.9 \\
  \cmidrule{2-7}
  & HumanEval & Acc         &  34.8 & -17.7 & -23.8 & -34.8 \\
  &           & Ret.        &  81.5 & +14.8 & +16.7 & +9.5 \\
\bottomrule
\end{tabular}
\caption{Evaluation results on Llama3-8B and Mistral-v0.3-7B, with performance accuracy in problem solving (Acc.) and retrieval scores measured by Exact Match (Ret.). The results at 0-token show the absolute value of the baseline performance under the original datasets. The later columns present the differences (deltas) between performances under different input lengths, where each problem is extended with the corresponding numbers of PaulGrahamEssay tokens  \textit{between} \texttt{evidence} and \texttt{question}(illustrated as \cref{fig:filler_experiment}), and those under 0-token. Scores in percentage.}

\label{tab:acc&retrieval}
\end{table*}

\begin{table*}[htp]
\centering 

\begin{tabular}{@{} llrrrrr @{}}
\toprule
Model & Task & 0 & 3750 & 7500 & 15000 & 30000 \\
\midrule
\multirow{4}{*}{Llama}
 & VarSum    & 96.0 & -5.0  & -8.0  & -12.0 & -48.0 \\
 & GSM8K     & 87.5 & -0.4  & -4.7  &  -3.2 &  -7.0 \\
 & MMLU      & 63.2 & -11.5 & -15.9 & -15.3 & -20.2 \\
 & HumanEval & 57.3 & -10.4 & -12.2 & -12.8 & -31.7 \\
\midrule
\multirow{4}{*}{Mistral}
 & VarSum    & 68.0 &  -3.0 &  -4.0 & -17.0 & -28.0 \\
 & GSM8K     & 70.0 & -20.0 & -12.0 & -24.0 & -30.0 \\
 & MMLU      & 54.1 &  -5.7 &  -9.2 & -12.7 & -14.0 \\
 & HumanEval & 34.8 &  -4.9 &  -4.9 &  -6.7 & -11.0 \\
\bottomrule
\end{tabular}
\caption{Performance drop across different context lengths on Llama-3-8B Instruct and Mistral-v0.3-7B-Instruct, with corresponding numbers of whitespace tokens inserted \textit{between} \texttt{evidence} and \texttt{question}. Values at context length > 0 are differences with 0-context.}  
\label{tab:space-filler}
\end{table*}

\begin{table*}[htp]
\centering 
\begin{tabular}{llrrrrrrrr}
\toprule
Model & Task & 0 & 7500 & 11250 & 15000 & 18750 & 22500 & 26250 & 30000 \\
\midrule
\multirow{4}{*}{Mistral}
  & Var Sum     & 66.0  & -4.0  & -7.0  & -7.0  & +3.0  & -3.0  & -15.0 & -17.0 \\
  & GSM8K       & 64.2  & -7.9  & -7.3  & -9.3  & -8.7  & -11.0 & -12.2 & -7.9  \\
  & HumanEval   & 34.8  & -6.7  & -6.1  & -6.7  & -5.5  & -6.1  & -8.0  & -9.8  \\
  & MMLU        & 54.0  & -5.5  & -7.2  & -7.2  & -7.4  & -9.2  & -9.5  & -10.1 \\
\midrule
\multirow{4}{*}{Llama3}
  & Var Sum     & 97.0  & -8.0  & -6.0  & -6.0  & -10.0 & -15.0 & -16.0 & -20.0 \\
  & GSM8K       & 85.5  & -5.6  & -6.6  & -8.1  & -10.7 & -5.6  & -8.4  & -7.1  \\
  & HumanEval   & 57.3  & -7.9  & -7.9  & -12.2 & -7.9  & -12.8 & -14.6 & -18.3 \\
  & MMLU        & 62.9  & -12.1 & -12.9 & -15.6 & -16.6 & -18.2 & -18.7 & -19.4 \\
\bottomrule
\end{tabular}
\caption{Performance across different context lengths on Llama-3-8B Instruct and Mistral-v0.3-7B-Instruct, with whitespace tokens of corresponding lengths inserted \textit{before} \texttt{evidence}, and \texttt{question} adjacent to \texttt{evidence}. Values at context length > 0 are deltas from the 0-context baseline.}
\label{tab:whitespace}
\end{table*}

\begin{table*}[htbp]
\centering  
\begin{tabular}{llrrrrrrrrr}
\toprule
\multirow{2}{*}{Model} & \multirow{2}{*}{Task} & 
\multicolumn{9}{c}{Context Length (tokens)} \\ \cmidrule(lr){3-11}
& & 0 & 3750 & 7500 & 11250 & 15000 & 18750 & 22500 & 26250 & 30000 \\
\midrule
\multirow{3}{*}{Llama3}
  & VarSum & 97.0 & -11.0 & -35.0 & -27.0 & -24.0 & -36.0 & -48.0 & -70.0 & -50.0 \\
  & GSM8K  & 86.1 & -1.7  & -3.3  & -6.9  & -4.3  & -9.4  & -11.7 & -18.0 & -19.6 \\
  & MMLU   & 62.8 & -11.3 & -15.9 & -16.6 & -15.5 & -19.0 & -18.1 & -18.7 & -21.1 \\
\midrule
\multirow{3}{*}{Mistral}
  & VarSum & 66.0 & -5.0  & -11.0 & -14.0 & -19.0 & -26.0 & -15.0 & -23.0 & -34.0 \\
  & GSM8K  & 64.5 & -2.1  & -4.8  & -6.5  & -8.2  & -11.6 & -11.6 & -14.4 & -15.1 \\
  & MMLU   & 53.8 & -4.7  & -7.5  & -8.8  & -11.0 & -10.9 & -11.5 & -9.9  & -11.8 \\
\bottomrule
\end{tabular}
\caption{Masking}
\label{tab:masking-filler-acc-ret}
\end{table*}

\subsection{Scientific Artifacts Used in this Work}

The datasets used in the work are presented in \cref{tab:artifacts}. This work only uses the artifacts for research purposes and does not redistribute them. The datasets do not contain sensitive or offensive content.

\newcolumntype{L}{>{\centering\arraybackslash}m}
\begin{table*}[htp]
\centering 
\begin{tabular}{L{.19\linewidth}L{.14\linewidth}L{.15\linewidth}L{.1\linewidth}L{.18\linewidth}}
\toprule
Dataset & Domain & Size & Language & License  \\
\midrule
GSM8K \cite{cobbe2021trainingverifierssolvemath}
 & Math & 8.79K Samples & English & MIT License \\
MMLU \cite{hendrycks2021measuringmassivemultitasklanguage} & QA & 116K Samples & English & MIT License \\
HumanEval \cite{chen2021codex} & Coding & 164 Samples & English; Python & MIT License \\
RULER \cite{hsieh2024ruler} & Long-context; 
Multitasking & 13 Tasks & English & Apache License 2.0 \\
\bottomrule
\end{tabular}
\caption{Artifacts used in this work}
\label{tab:artifacts}
\end{table*}

\subsection{Computational Resources and Experiment Statistics of this Work}

This work uses GH200 GPUs for computations. Around 20000 GPU hours are consumed. All the experiments are run once, the results of which are reported. 

\subsection{Potential Risks}

This work explores the phenomenon of performance degradation in long-context scenarios, which is in itself unlikely to cause potential risks. This work also presents a strategy to mitigate the issue, which is mainly for demonstration purposes, and its reliability is not guaranteed.

\end{document}